# The Ethical Implications of Shared Medical Decision Making without Providing Adequate Computational Support to the Care Provider and to the Patient

Yuval Shahar, M.D., Ph.D., FACMI, MIAHSI

Medical Informatics Research Center
Department of Software and Information Systems Engineering
Ben Gurion University,
Beer Sheva 84105,
Israel
Email: yshahar@bgu.ac.il

**Abstract**

There is a clear need to involve patients in medical decisions. However, cognitive psychological research has highlighted the cognitive limitations of humans with respect to (a) probabilistic assessment of the patient's state and of potential outcomes of various decisions, (b) elicitation of the patient's utility function, and (c) integration of the probabilistic knowledge and of patient preferences to determine the optimal strategy.

Therefore, without adequate *computational* support, current shared-decision models have severe ethical deficiencies. An "informed-consent" model unfairly transfers the responsibility to a patient who does not have the necessary knowledge, nor the integration capability. A "paternalistic" model endows with exaggerated power a physician who might not be aware of the patient's preferences, is prone to multiple cognitive biases, and whose computational integration capability is bounded.

Recent progress in Artificial Intelligence suggests adding a *third* agent: a *computer*, in all *deliberative medical decisions*: Non-emergency medical decisions in which (a) more than one alternative exists, (b) the patient's preferences can be elicited, (c) the therapeutic alternatives might be influenced by these preferences, (d) medical knowledge exists regarding the likelihood of the decisions' outcomes, and (e) there is sufficient decision time.

Ethical physicians should exploit computational decision-support technologies, neither making the decisions solely on their own, nor shirking their duty and shifting the responsibility to patients in the name of "informed consent." The resulting three-way (patient, care provider, computer) human-machine model that we suggest emphasizes the patient's preferences, the physician's knowledge, and the computational integration of both aspects, does not diminish the physician's role, but rather brings out the best in human and machine.

# The ethics of shared medical decision making without computational support

## Introduction: Classical and modern physician-patient interaction models

Several models had been proposed for the delicate process underlying the physician-patient encounter, leading to multiple worthwhile insights regarding the best way to involve patients in the therapeutic process. Clearly, the paternalistic model in which the physician makes all of the decisions is not sufficient, for multiple functional and ethical reasons.

Thus, significant emphasis has been put on the need for *shared decision-making* and on its ramifications, such as *informed consent*. However, informed consent is not a panacea: Braddock et al.[1] analyzed 1057 taped patient encounters with 59 primary care physicians and 65 surgeons; only 9% of the encounters met the authors' criteria for completeness and informed decision making (e.g., a discussion of the nature of the intervention). Nevertheless, the professional, legal, and moral necessity for informed consent, in spite of potential patient-physician communication problems, has been persuasively argued by Doyal[2], who points out that although certain errors of risk evaluation do plague patients, not all patients make them, and many of them can benefit from information customized to their needs; and that although many patients would leave the decisions in the hands of their physicians, many others do desire information about their choices and can make them coherently and to good effect. Woolf et al.[3] listed three options of offering decision counseling: By clinicians who lack any formal training, by clinicians with informed-choice training, and by impartial decision councilors. None appears ideal, and controlled studies are needed to determine which is best.

Miller[4] has analyzed more abstractly what constitutes an *autonomous* patient decision by listing four different senses of autonomy: *Free action* (voluntary and intentional); *authentic choice* (a choice consistent with previous values and plans of the patient); an *effective deliberation* process (the patient is aware of the need to decide and of the alternatives offered, and evaluates all options); and *moral reflection* (the patient consciously and critically examines her values and decides whether to continue to live with them.) Miller argued that if a patient is choosing without authenticity or effective deliberation, her physician has a stronger obligation to question her decisions and enhance her autonomous capacities before making the final decision. Considering the *mode* of the physician-patient interaction, Emanuel and Emanuel's[5] influential paper considered four models for that interaction. After rejecting the *paternalistic*, *informative*, and *interpretive* models, the authors focus on a *deliberative* model, in which patient and physician engage in deliberation about the most worthwhile health-related values, and the best course of action to pursue these values.

However, as we shall now discuss, the current physician-patient interaction models do not seem to sufficiently take into account the developments of the past several decades in cognitive psychology, computer science, and artificial intelligence, and the serious ethical implications of ignoring these findings.

**The ethics of shared medical decision making without computational support**

**The problems inherent in the various physician-patient interaction models, in the light of current psychological and computational research**

The various interaction models discussed in the medical and philosophical literature represent well the state of the art during much of the mid 20th Century, and in some cases, up to its end. However, it seems that previous authors might not have given sufficient weight to the following four considerations, which have far-reaching implications as to what constitutes a *truly* autonomous patient decision or a *meaningful* deliberative process, and as to what agent and/or process would best represent the patient's preferences. Thus, we believe that the current discussion of the binary patient-physician relation might have missed a *third* agent that, at least in the 21st century, should be added to it; an agent that has already been demonstrated to have a significant potential for enhancing, in an objective manner, the quality and the ethical aspects of the clinical decision-making process.

The four considerations are the following:

1. The past four decades of cognitive-psychology research have rigorously documented the inherent judgmental biases of humans (both patients and physicians). A major example is Tversky and Kahneman's seminal investigation of three basic categories of heuristics (*representativeness*, *availability*, *anchoring & adjustment*) and the biases that they imply, which are inherent in the way humans estimate *probabilities*[6]. These biases affect the decisions based on the probability estimates, since most decisions require appropriate consideration of the prior probability of the disease, of the sensitivity and specificity of the multiple diagnostic tests now available, and of the effects of each potential course of action on the patient's state. Behavioral scientists and economic researchers have repeatedly validated these insights; researchers such as David Eddie have even demonstrated that physicians seem to confuse the *sensitivity* of the diagnostic test to the disease, with its positive-predictive value, i.e., the probability of the disease given a positive result of the test, providing an estimate for the probability of a malignancy that is a whole order of magnitude larger than the true one[7]. Unfortunately, Poses, Cebul, and Wigton[8] had demonstrated that education is *not* a sufficient remedy: Teaching physicians how to improve their estimate of disease probabilities in the domain of streptococcal pharyngitis did *not* effect their treatment decisions.

2. The research into the biases inherent in the process of eliciting the patient's *preferences* (i.e., their context-sensitive utility function) has cast serious doubts on whether physicians can, without any training or external human or computational support, correctly elicit these preferences. However, optimal decision making requires both correct assessment of probabilities and effective elicitation of utility functions.

    There are in fact two opposing aspects to this consideration:

# The ethics of shared medical decision making without computational support

   a. On one hand, there is an inherent bias during the utility-elicitation process, also investigated by Tversky and Kahneman, in the manner of framing the decisions in which the patient's preferences are to be used[9]. In a classical study, McNeil et al. have highlighted the effect of framing on the choices of patients and physicians[10], while Meyerowitz and Chaiken[11] have demonstrated the effect of positive and negative framing on persuading women to perform breast self-examination. Knowing the patient's preferences and the relevant probability of success of various potential therapeutic alternatives is crucial for determining a patient-specific optimal therapy. Common clinical thinking often puts the main emphasis on longevity. However, as demonstrated by McNeil et al., by actually asking lung-cancer patients who need to decide whether to undergo risky surgery, versus irradiation, whether 5-year survival is their main goal, it is clear that at least for a subset of the patients, this was not the case[12]. Unfortunately, given the physician-patient power-differential aspect, the patient is often at a serious disadvantage, and, as has previously been pointed out persuasively by Rubin, the "freedom to choose" may well be the "inability to refuse[13]".

   b. On the other hand, fortunately, multiple studies had demonstrated the feasibility of using various techniques, often implemented by computational tools, for elicitation of patient preferences in various domains[14, 15]. Examples range from treatment of mental-health disorders while avoiding the side effects of medications, through prenatal testing[16] and postmenopausal hormone therapy, to the therapy of deep vein thrombosis[17]. In the case of cancer patients, using a self-assessment tool led to greater congruence between the patients' self-reported problems and preferences and those addressed by their health care providers[18].

3. Medical decisions are, in fact, *strategies* for maximization of the patient's (personalized) expected utility, which often need to consider multiple complex factors and future implications, such as complications and benefits, which in turn depend on the probabilistic assessment of the patient's initial state, as well as on the probabilistic assessment of the future effects of each potential course of action, and require extended reasoning into the future. However, human computational capabilities, whether those of physicians or of their patients, are limited; the human (in)ability to correctly compute the optimal strategy that maximizes the patient's expected utility is sometimes referred to (following psychologist, economist, and pioneer artificial-intelligence researcher, Herbert Simon) as *bounded rationality*. In particular, even if physicians could correctly assess all relevant probabilities, and even if they could accurately elicit all of the patient's multi-faceted utility functions, it is highly doubtful whether they could correctly integrate these preferences and probabilities when determining the best strategy to maximize the patient's benefit.

# The ethics of shared medical decision making without computational support

For example, it is important to consider explicitly the trade-off between longevity and quality of life, as was powerfully highlighted by McNeil et al., in the case of surgery versus irradiation for laryngeal cancer: A significant subset of patients would have preferred irradiation over surgery, even if it leads to a slightly decreased 5-year survival, due to its beneficial effect of retaining their capability for speech[19]. That, however, was not the treatment of choice originally offered to these patients, nor was it compared to the treatment of choice, while considering each patient's preferences.

4. The potential of computational decision-support systems that partially or fully use decision-theoretic models is being increasingly demonstrated and recognized. These emerging systems are potentially capable of integrating patient data, patient preferences, and (probabilistic) relevant evidence-based medical knowledge, to provide a patient-specific recommendation, or at least a rational for diagnostic information gathering and assessment, in manner that Heckerman et al. had already demonstrated even in the early days of the currently exploding with innovation area of *Artificial Intelligence in Medicine*[20]. Furthermore, decision-support architectures had been developed that facilitate shared decision making, by providing effective tools for elicitation of the patient's preferences[21]. In addition, medical decision-support systems are often capable of providing a multi-dimensional *sensitivity analysis* that *explains* the factors affecting a recommended strategy and that increases the robustness of the patient's decision[22]. Other types of computational decision-support systems have been shown to increase the accuracy of assessment of patient state following bone-marrow transplantation from 57% to 92%[23]; to improve the outcome of brain stroke by enhancing compliance to established clinical guidelines, which directly affects the clinical outcomes[24]; and to assist general practitioners, residents, and even specially-trained nurses in the administration of guideline-based care in multiple clinical domains, such as pre-eclampsia/toxemia of pregnancy, and increase the compliance with the guideline's recommended actions from 49% to 93%[25,26]. Indeed, although research in the area of automated support to guideline-based care requires additional evaluation, the technology for conversion of most evidence-based textual clinical guidelines into machine-comprehensible representations, which can be applied by computers to assist in the management of patients, already exists[27]. This technology is especially important for chronic-disease patients, whose management requires significant resources for continuous monitoring and therapy. In fact, a 13-partners, 5-countries EU academic-industrial research project, *MobiGuide*, has integrated many of these evidence-based clinical decision-support tools, to provide continuous management of chronic patients at home, through sensors on the patients and smart phones, providing the patients with continuous alerts and personalized guideline-based recommendations, and, in parallel, the patients' care providers with evidence-based decision support regarding these patients[28,29]. Clinical decision-support systems have become an established area, in which successes and failures are objectively discussed[30].

# The ethics of shared medical decision making without computational support

Finally, modern Artificial Intelligence systems, and in particular, *Deep Learning* architectures[31], a technology for which a Turing Award was recently bestowed in the Computer Science field, have a significant potential for enhancing the physician's probabilistic assessment and interpretation of the patient's state and for reducing human errors, through these systems' high and consistent diagnostic level. Modern Deep Learning systems often reach an FDA-approved status or expert-physician level, such as for autonomous diagnosis of diabetic retinopathy when provided with relatively low-quality fundus photographs[32,33], classification of various skin lesions, and in particular, skin cancer, at least at the accuracy of a dermatologist, when given simple dermatological images[34], or detection of metastases in breast cancer, at the level of an expert pathologist with unlimited time, when given access to digital-pathology lymph-node slides[35].

The first three considerations strongly suggest that humans, including physicians, let alone seriously ill patients, cannot be expected to correctly perform the difficult computational task of correctly assessing and computing the relevant probabilities of disease states and potential outcomes, eliciting the relevant patient-specific preferences, and, in particular, integrating both into a coherent patient-specific strategy.

The fourth consideration suggests a viable alternative: Assistance by adding to the patient-physician diad a *third agent*, a *computational agent*, thus leading to a patient-physician-computer triad. Probabilistic computation *necessitates* computational assistance; preference elicitation could benefit from it; and overall effective decision making desperately requires it.

Thus, we argue here that a truly *authentic* patient choice, in Miller's sense, is practically impossible without *major computational assistance,* which, the fourth consideration suggests, is potentially quite feasible in the new, triad-based architecture. Only a rational computational agent whose recommendations can be explicitly examined (by physician and patient alike) and can be explained or justified by known evidence, estimated prior probabilities, and patient values, can cater for all of our implied desiderata, at least in deliberative domains in which the diagnostic or therapeutic alternatives are not straightforward and might potentially be influenced by patient preferences.

## The Ethical Implications of Bounded Human Rationality to the Concept of Shared Medical Decision Making

The serious *ethical* problems implied by the current situation are now apparent.

First, the older, paternalistic view of the patient, leaving all decisions to the clinician, is *not* a viable alternative we can go back to: Such an archaic view ignores the patient's preferences altogether, while introducing the physician's own probabilistic, preferential, and computational biases. Given the well-known inherent human psychological framing biases, the potential conflicts of interest involved in any physician-patient relationship, and the inherent probabilistic nature of most of the clinical contexts relevant to a deliberative informed consent, there are serious grounds to doubt that physicians can correctly acquire the patient's preferences, AND correctly compute the implications of these

# The ethics of shared medical decision making without computational support

preferences in the light of probabilistic information, such as the results of certain diagnostic tests and the known accuracy of these tests.

Second, ignoring the psychological and computational aspects of the tasks of assessment and integration of probabilistic clinical knowledge with the patient's preferences, and leaving patients to their own resources in the name of "informed consent", is even worse. It is tantamount, in our opinion, to literally *shirking a part of the physician's responsibility for the patient*. We suspect that such an approach is often motivated more by the legal preferences of the clinicians or of the institutions in which they are employed, than by the clinical preferences of the patient. Thus, the concept of "informed consent" without adequate computational support to the physician and, especially, to the patient, is somewhat of a misnomer; *in reality, the patient is often neither fully informed, nor can she provide a meaningful consent, in the sense of fully recognizing the implications of her choice.*

Thus, a serious ethical concern should be raised as to the validity of *any* model involving *manual* (as opposed to a computationally aided) assessment of probabilities, or computation of optimal [for the patient] strategies, by either patients or care providers. *Such a stance is not unlike refusing to use a stethoscope or a chest X-ray, preferring instead to listen to the patient's heart and lungs through their clothes.*

Of course, this analysis is only relevant to appropriate clinical contexts, which we can refer to as *deliberative medical domains*: non-emergency domains in which more than one alternative exists, it is possible to elicit the patient's values, the therapeutic alternatives might potentially be influenced by patient preferences, the medical knowledge exists regarding the likelihood of different outcomes given each alternative, and there is sufficient time for deliberative decision making. Fortunately, however, multiple clinical domains, from anti hypertensive therapy to genetic consultation, answer these requirements, and significant experience exists with respect to assessing patient preferences in them to a positive effect.

**The Solution: The Clinical Holy Triad, or: The Computer is the Patient's Best Friend**

In our view, a truly *ethical* physician must strive to provide her patient with the very best advice, customized to the patient's data and preferences. Provision of such an advice can be significantly assisted by the use of an appropriate computational agent, namely, the modern computer in its various reincarnations, which we believe is the missing element in what should be a clinical *triad*, in addition to the patient-physician diad.

Often, *an objective computational agent* can assist (1) in retrieving, assessing, and interpreting the inherently uncertain patient-specific data, especially in the light of current medical knowledge, (2) in eliciting the patient's preferences, and, in particular, (3) in providing (and critically examining and explaining) the best recommendation based on these patient-specific probabilistic assessments, and on the patient's preferences.

# The ethics of shared medical decision making without computational support

Thus, given such an objective computational assistant, and an appropriate clinical domain, the physician's important roles, viewed through our shared decision-making lenses, consist mainly of determining the patient's overall state and context (e.g., laryngeal cancer at a particular stage), determining the relevant options (e.g., surgery, irradiation), selecting the appropriate clinical decision model (e.g., the "*laryngeal cancer*" module, which might include the currently up-to-date probabilistic estimates of the effects of each course of action), overseeing a manual or semi-automated *elicitation* of the patient's preferences (including the validation of these preferences), providing the *computational* model with the necessary patient-specific data and elicited preferences, *interpreting* the results to the patient, and (if needed) assisting in the process of *sensitivity analysis*, by modifying both relevant clinical information and patient values to see how stable the patient's decision is, thus adding iteration and feedback to the process. We refer to this model as the *elicitation, computation, interpretation*, and *sensitivity analysis* [ECIS] model, or as the *computational consultant* model.

The ECIS model, whenever it can be effectively applied (a situation that is becoming increasingly feasible, given the current Artificial Intelligence technology) addresses all of our core ethical concerns. Neither patient nor physician is faced with the humanly impossible tasks of correctly computing all of the relevant probabilities, or the optimal decisions based on their integration with the patient's preferences. Support, using the best available computational tools, can also be provided to the difficult task of the very elicitation of relevant preferences. The integration of these preferences with the known clinical probabilistic evidence (including the most up-to-date genetic findings) can provide a rational basis for recommendation of an optimal personalized therapeutic strategy, customized to the patient's genotype, clinical phenotype, and personal preferences. Decision robustness can then be provided through a sensitivity analysis of the effect of modifying the estimated probabilities and/or the patient's preferences[21]. Thus, the ECIS model includes the important aspects of Emanuel and Emanuel's *interpretive* and *deliberative* encounter models, while leaving the computational tasks to an objective, rational agent, who can also assist the provider and/or the patient in the preference-elicitation task.

The ECIS model, which introduces the computer as a third, vital aspect of a patient-physician-computational agent triad, in no way diminishes the physician's important role[s], but rather brings out the best in human and machine, and, in our view, might often provide the most *ethical*, truly shared decision-making process, without imposing on either the patient or the physician computational tasks that neither is sufficiently equipped to handle.

## Acknowledgments

We would like to acknowledge the multiple authors' whose studies are mentioned in the References, for providing us with valuable insights regarding Shared Decision Making, sometimes through reading their papers, often through actual discussions with the authors.

## References

1. Braddock CH., Edwards KA., Hasenberg N, et al. Informed decision making in outpatient practice: Time to get back to basics. *J Am Med Inform Assoc* 1999; **282**(24):2313-2320.

# The ethics of shared medical decision making without computational support

2. Doyal L. Informed consent: Moral necessity or illusion? *Qual Health Care* 2001;**10**(Supplement i):i29-i33.

3. Woolf SH, Chan ECH, Harris R, et al. Promoting informed choice: transforming health care to dispense knowledge for decision making. *Ann Intern Med* 2005;**143**(4):293-300.

4. Miller BL. Autonomy and the Refusal of Lifesaving Treatment. *Hastings Center Report* 1981; **11**(4): 22-28.

5. Emanuel EJ, Emanuel LL. Four Models of the Physician-Patient Relationship. *JAMA* 1992;**267** (16): 221-6.

6. Tversky A, Kahneman D. Judgment under uncertainty: Heuristics and biases. *Science* 1974;**185**:1124-1130.

7. Eddy, D. Probabilistic reasoning in clinical medicine: Problems and opportunities. In: Kahneman D., Slovic P., and Tversky A. (eds). *Judgment under uncertainty: Heuristics and biases*. Cambridge, England: Cambridge university press 1982: 249–267.

8. Poses RM, Cebul CD, Wigton RS. You can lead a horse to water-improving physician's knowledge of probabilities may not affect their decisions. *Med Decis Making* 1995;**15**:65-75.

9. Tversky A, Kahneman D. The framing of decisions and the psychology of choice. *Science* 1981;**211**:453-458.

10. McNeil BJ, Pauker S, Sox Jr. H, et al.. On the elicitation of preferences for alternative therapies. *N Engl J Med* 1982;**306**:1259-1262.

11. Meyerowitz BE, Chaiken S. The effect of message framing on breast self examination attitudes, intentions, and behavior. *J Pers Soc Psychol* 1987; **52**:500-520.

12. McNeil BJ, Weichselbaum R, Pauker SG. Fallacy of the five-year survival in lung cancer. New *N Engl J Med* 1978; **299**(25):1397-1401.

13. Rubin SS. The multiple roles and relationships of ethical psychotherapy: Revisiting the ideal, the real, and the unethical. In: Lazarus A. and Zur O. *Dual Relationships*. New York, NY: Springer 2002.

14. Goldstein MK and Tsevat J. Applying utility assessment at the bedside. In: Chapman, G.B. and Sonenberg, F.A., eds. *Decision Making in Health Care*. Cambridge, England: Cambridge university press 2000.

15. O'Connor AM, Rostrom A, Fiset V, et al. Decision aids for patients facing health treatment or screening decisions: systematic review. *BMJ* 1999;**319**:731-735.

16. Kuppermann M, Feeny D., Gates E, et al. Preferences of women facing a prenatal diagnostic choice: Long-term outcomes matter most. *Prenat Diagn* 1999;**19**:711-716.

17. Lenert LA, Soetikno RM. Automated computer interviews to elicit utilities: potential applications in the treatment of deep venous thrombosis. *J Am Med Inform Assoc* 1997;**4**:49-56.

18. Ruland CM, White T, Stevens M, Fanciullo G, Khilani S. Effects of a computerized system to support shared decision making in symptom management of cancer patients: Preliminary results. *J Am Med Inform Assoc* 2003;**10**(6):573-579.

19. McNeil BJ, Weichselbaum R, Pauker SG. Speech and survival: tradeoffs between quality and quantity of life in laryngeal cancer. *N Engl J Med* 1981;**305**(17):982-987.

20. Heckerman DE, Horvitz EJ, and Nathwani, BN.   Towards normative expert systems: Part I.  The Pathfinder project.  *Methods Inf Med* 1992;**31**:90-105.

21. Quaglini, S., Miksch, S., Shahar, Y., Peleg, M., Napolitano, C. Rigla, M., Pallàs, A., Enea Parimbelli, E., and Sacchi, L. Supporting shared decision making within the MobiGuide project. *Proceedings of the 2013 American Medical Informatics Association Annual Fall Symposium (AMIA-2013)*, 2013, Washington, DC.

22. Segal I, Shahar Y. A distributed shared medical decision-making framework for recommendation, exploration, and explanation of clinical strategies.  *J Biomed Inform* 2009;42(2):272-286.

23. Martins S.B., Shahar Y., Goren-Bar D., Galperin M., Kaizer H., Basso L.V., McNaughton, D., and Goldstein, M.K. Evaluation of an architecture for intelligent query and exploration of time-oriented clinical data. *Artif Intell Med* 2008;43:17-34.

24. Micieli, G., Cavallini, A., Quaglini., S.   Guideline Compliance Improves Stroke Outcome – A Preliminary Study in 4 Districts in the Italian Region of Lombardia.   *Stroke* 2002;33:1341-1347.

25. Shalom, E., Shahar, Y., and Lunenfeld, E. An architecture for continuous, user-driven, and data-driven application of clinical guidelines: addressing the realistic aspects of clinical decision support. *The Journal of Biomedical Informatics* 2016;**59**: 130-148.

26. Shalom, E., Shahar, Y., Parmet, Y., and Lunenfeld, E. A multiple-scenario assessment of the effect of a continuous-care, guideline-based decision support system on clinicians' compliance to clinical guidelines. *The International Journal of Medical Informatics* 2015;84 (4):248-262.

27. Latoszek-Berendsen, A., Tange, H., van den Herik, H.J., Hasman, A.: From clinical practice GLs to computer-interpretable GLs. A literature overview.  *Methods Inf Med* 2010;**49**(6):550-70.

28. Peleg, M., Shahar, Y., Quaglini, S., Fux, A., Garcia-Sanchez, G., Goldstein, A., González-Ferrer, A., Hernando, Jones, V., Klebanov, G., M.H., Klimov, D., Broens, T., Knoppel, D., Larburu, N., Marcos, C., Martinez-Sarriegui, I., Napolitano, C., Pallas, A., Palomares, A., Parimbelli, E., Rigla, M., Sacchi, L., Shalom, E., Soffer, P., and van Schooten, B.  Assessment of a personalized and distributed patient guidance system. *The International Journal of Medical* Informatics 2017;**101**:108-130.

29. Peleg, M., Shahar, Y., Quaglini, S., Fux, A., Garcia-Sanchez, G., Goldstein, A., Hernando, M.H., Klimov, D., Martinez-Sarriegui, I., Napolitano, C., Rigla, M., Sacchi, L., Shalom, E., and Soffer, P. MobiGuide: a personalized and patient-centric decision-support system and its evaluation in the atrial fibrillation and gestational diabetes domains. *User Modeling and User Adapted Interaction* 2017;**27**(2):159–213.

30. Greenes, R.A., Bates, D.W., Kawamoto, K., Middleton, B., Osheroff, J., and Shahar, Y.  Clinical decision support models and frameworks:  seeking to address research issues underlying implementation successes and failures. *The Journal of Biomedical Informatics* 2018;**78**:134-143.

31. LeCun, Y., Bengio, Y. & Hinton, G. Deep learning. *Nature* 2015;**521,** 436–444.

32. Abràmoff, M.D., Lavin, P.T., Birch, M. *et al.* Pivotal trial of an autonomous AI-based diagnostic system for detection of diabetic retinopathy in primary care offices. *npj Digital Med* 2018; **1,** 39.

33. Gulshan V, Peng L, Coram M, et al. Development and validation of a deep learning algorithm for detection of diabetic retinopathy in retinal fundus hotographs. *JAMA.* 2016;**316**(22):2402–2410.

34. Esteva, A., Kuprel, B., Novoa, R. *et al.* Dermatologist-level classification of skin cancer with deep neural networks. *Nature* 2017; **542,** 115–118.

35. Ehteshami Bejnordi B, Veta M, Johannes van Diest P, et al. Diagnostic assessment of deep learning algorithms for detection of lymph node metastases in women with breast cancer. *JAMA.* 2017;**318**(22):2199–2210.